\documentclass[runningheads]{llncs}

\usepackage{eccv}

\usepackage{eccvabbrv}

\usepackage{graphicx}
\usepackage{booktabs}

\usepackage[accsupp]{axessibility}  
\usepackage{multirow}
\usepackage{siunitx}  

\usepackage[pagebackref,breaklinks,colorlinks,citecolor=eccvblue]{hyperref}
\usepackage[table]{xcolor}
\usepackage{orcidlink}

\begin{document}

\title{P$^{3}$Nav: End-to-End Perception, Prediction and Planning for Vision-and-Language Navigation} 

\titlerunning{P$^{3}$Nav}


\author{Tianfu Li\inst{1}\textsuperscript{*}\orcidlink{0009-0007-2410-9595} \and
        Wenbo Chen\inst{1}\textsuperscript{*}\orcidlink{0009-0001-9323-0555} \and
        Haoxuan Xu\inst{1}\textsuperscript{*}\orcidlink{0009-0006-6435-8900} \and \\
        Xinhu Zheng\inst{1}\orcidlink{0000-0002-9898-5543} \and
        Haoang Li\inst{1}\textsuperscript{\dag}\orcidlink{}}

\authorrunning{T.~Li et al.}

\institute{HKUST(GZ), Guangzhou, China \\
}

\maketitle


\renewcommand{\thefootnote}{}
\footnotetext{\textsuperscript{*}Equal contribution, \textsuperscript{\dag}Corresponding author}

\begin{abstract}
  In Vision-and-Language Navigation (VLN), an agent is required to plan a path to the target specified by the language instruction, using its visual observations. Consequently, prevailing VLN methods primarily focus on building powerful planners through visual-textual alignment. However, these approaches often bypass the imperative of comprehensive scene understanding prior to planning, leaving the agent with insufficient perception or prediction capabilities. Thus, we propose P$^{3}$Nav, a novel end-to-end framework integrating perception, prediction, and planning in a unified pipeline to strengthen the VLN agent’s scene understanding and boost navigation success. Specifically, P$^{3}$Nav augments perception by extracting complementary cues from object-level and map-level perspectives. Subsequently, our P$^{3}$Nav predicts waypoints to model the agent's potential future states, endowing the agent with intrinsic awareness of candidate positions during navigation. Conditioned on these future waypoints, P$^{3}$Nav further forecasts semantic map cues, enabling proactive planning and reducing the strict reliance on purely historical context. Integrating these perceptual and predictive cues, a holistic planning module finally carries out the VLN tasks. Extensive experiments demonstrate that our P$^{3}$Nav achieves new state-of-the-art performance on the REVERIE, R2R-CE, and RxR-CE benchmarks.
\end{abstract}

\section{Introduction}
\label{sec:intro}

\begin{figure}[tb]
  \centering
  \includegraphics[width=1\linewidth]{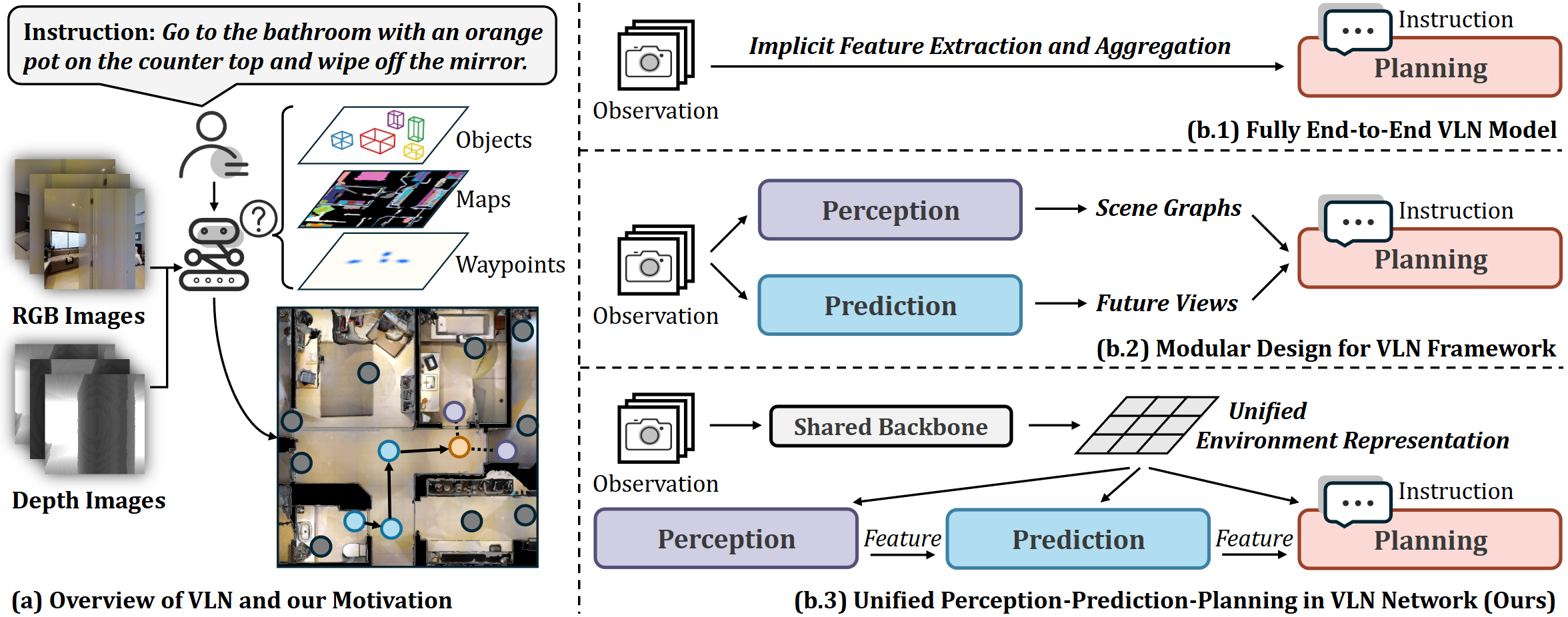}
  \caption{Motivation of our method. (a) Scene information, such as objects, maps, and waypoints are crucial for grounding instructions to visual observations. 
  (b.1) Early models are of a ``planning-only'' structure, with limited scene understanding due to implicit feature extraction and aggregation. 
  (b.2) Recent methods build external perception/prediction modules, but suffer from information loss and error accumulation.
  (b.3) Our method unifies perception, prediction, and planning in a single network, through a unified environment representation with end-to-end feature propagation between each stage.}
  \label{fig:overview}
\end{figure}

Enabling agents to understand natural language instructions and autonomously execute tasks is an indispensable step in embodied AI. 
Against this backdrop, Vision-and-Language Navigation (VLN) has drawn surging attention~\cite{R2R,VLN-PE,song2025towards}, which requires an agent to find a language-specified target, using its visual observations~\cite{ETPNav, BEVBert,SAME}, as shown in~\Cref{fig:overview}(a). 
Consequently, the main VLN methods focus on building more powerful planning modules by enhancing visual-textual association learning~\cite{BEVBert, HOP, HOP2,MTU3D}, and substantial progress has been made.

However, existing fully end-to-end VLN models rely on implicit feature extraction and aggregation~\cite{BEVBert, GridMM} (shown in~\Cref{fig:overview}(b.1)), lacking explicit guidance to capture navigation-critical scene information.
Subsequent works incorporate perception tasks as pre-training objectives to learn environment-aware backbones~\cite{BEVSceneGraph, VER}.
Yet, at inference time, these extracted semantic entities and spatial relations are not directly routed to the navigation policy. Without explicit perception cues, the agent still struggles to reliably align in-scene entities with their linguistic descriptions, frequently resulting in misguided navigational decisions.
To address this problem, recent works leverage external perceivers to build scene graphs~\cite{SpatialNav, FSR-VLN} or standalone predictors to forecast future view information~\cite{HNR-VLN,NavQ}, as demonstrated in~\Cref{fig:overview}(b.2). 
Although these approaches enhance scene understanding, such modular designs suffer from information loss and cascading error accumulation across modules~\cite{ST-P3,UniAD, VAD}. 
For instance, a noisy scene graph will inevitably mislead the downstream navigation policy.
Moreover, the view-based prediction forces the agent to implicitly reconcile incomplete and duplicate observations across discrete panorama faces~\cite{BEVBert}. This cognitive burden injects spatial ambiguities into the cross-modal planning stage and degrades navigation accuracy.
These combined issues ultimately result in impaired planning performance.

To this end, we propose a novel end-to-end VLN framework (\textbf{P$^{3}$Nav}), which integrates \textbf{perception}, \textbf{prediction}, and \textbf{planning} into a unified differentiable pipeline (shown in~\Cref{fig:overview}(b.3)). 
Through a unified environment representation with end-to-end feature propagation between each stage, P$^{3}$Nav equips the planner with scene information that spans both past observations and future forecasts, yielding enhanced scene understanding, finer grained alignment with linguistic instructions, and ultimately a higher navigation success rate.

Specifically, building upon a shared backbone, we base all the modules on a unified bird's-eye-view (BEV) representation and formulate perception from two complementary levels: object level and map level.
Since objects often appear as landmarks in the instructions, for example, \textit{``walk to the \textbf{chair}, and then ...''}, explicitly modeling these items yields visual embeddings that can be readily aligned with their corresponding textual tokens.
Furthermore, these landmark objects rarely appear in isolation; the spatial relations among them impose an additional constraint. For instance, the instruction \textit{``find the \textbf{pillow} \textbf{on} the \textbf{sofa}''} requires the agent to choose the pillow supported by the sofa rather than another pillow that may rest on the bed. 
Object-level perception alone, which only yields disjoint embeddings, cannot intuitively decide on this support relation. 
Thus, we perform map-level perception to project the objects' spatial relations in a single and unified tensor, strengthening the alignment between these landmarks and their corresponding phrases in the instruction. Hence, the jointly enhanced scene understanding via objects and map information improves ultimate planning.

Beyond perceiving the present, we task the agent to predict both where it might move and what it will see once it arrives. 
For the former task, the agent forecasts its instruction-agnostic potential states to get intrinsic awareness of candidate positions; this module also helps to tackle the discrete-to-continuous transfer issue. 
Conditioned on the potential states, a lightweight decoder further predicts future scene features, endowing the planner with forward-looking cues.
By shifting from view-based predictions to a map-based alternative, we spare the planner the burden of aligning fragmented view embeddings and instead deliver a unified tensor upon which downstream computation can be performed directly.
Together, these two streams enable the agent with predictive scene understanding, which also contributes to cross-modal planning.

Finally, the planning module takes the aforementioned information to make comprehensive navigation decisions. Crucially, instead of hard-decoding the intermediate tensors into human-readable physical symbols, we directly feed the penultimate features from each head that still reside in the shared metric coordinate frame to the planner. This design keeps the entire pipeline end-to-end differentiable, thereby reducing information loss, error accumulation, and negative transfer from isolated optimization targets~\cite{UniAD, PnPNet, Fast, Perceive}. 
P$^{3}$Nav establishes state-of-the-art (SOTA) records on REVERIE~\cite{RvR}, R2R-CE~\cite{BeyondNavGraph} and RxR-CE~\cite{RxR-CE} benchmarks.
To sum up, our P$^{3}$Nav contributes in four respects:

\begin{itemize}
    \item To the best of our knowledge, our P$^{3}$Nav is the first VLN model that unifies perception, prediction, and planning in a single network, using intermediate modules to sharpen scene understanding and boost planning accuracy.
    \item By modeling the landmark objects and their spatial relations via the dual-level perception mechanism, our P$^{3}$Nav enhances scene understanding and the visual-textual association in cross-modal planning.
    \item Leveraging the predicted future states of both the agent itself and the surrounding scene, our P$^{3}$Nav enables predictive understanding and planning. The map-based future scene features further relieve the planner from the burden of aligning discrete view-based embeddings.
    \item Our P$^{3}$Nav achieves SOTA performance on the REVERIE, R2R-CE and RxR-CE benchmarks.
\end{itemize}

\section{Related Work}

\noindent
\textbf{Vision-and-Language Navigation (VLN).}
VLN tasks an agent to navigate to the target location specified by language instructions~\cite{CLIP-Nav, lu2025monovln, zhang2025cosmo, wang2023dual, Wang2024GOAT}.
Early methods adopt sequence-to-sequence frameworks~\cite{Seq2Seq, R2R, SpeakerFollower} or transformer architectures~\cite{Transformer, VLNrBERT, VLNrBERT2, AirBERT, HAMT, xu2026enhancing} to directly map observations and instructions to actions. These approaches entrust perception capabilities entirely to the visual backbone, struggling to correlate discrete images across views~\cite{BEVBert}.
Subsequent research aggregates visual features into structured map representations, such as topological graphs~\cite{DUET, ETPNav, NavGPT2, MapGPT, LLMasCo, RxR-CE}, grid maps~\cite{GridMM}, or hybrid designs~\cite{BEVBert, BEVSceneGraph, VER}. Furthermore, some studies introduce auxiliary perception tasks as proxy pre-training objectives to enhance spatial awareness~\cite{BEVSceneGraph, VER, MapSupervision}. 
However, none of these methods incorporates perception results into the navigation policy during inference, leading to insufficient scene understanding.
To address this limitation, recent approaches adopt standalone perception and prediction modules. 
Specifically, the perceivers construct various scene graphs~\cite{FSR-VLN, SpatialNav, cao2025cognav} and metric maps~\cite{MapNav, cao2025cognav}; 
Moreover, the predictors forecast future RGB frames~\cite{DREAMWALKER} or view-based features~\cite{HNR-VLN, g3d, NavQ} through world models~\cite{li2025comprehensive}, neural radiance fields~\cite{NeRF} and Q-learning~\cite{Q-learning}.
However, such modular designs introduce information loss and error accumulation across modules~\cite{ST-P3,UniAD, VAD}. 
Consequently, our method unifies all modules in an end-to-end manner, feeding perception and prediction information into the planner to enhance navigation performance.

\noindent
\textbf{VLN in Continuous Environment (VLN-CE).}
Initially, VLN agents moved through pre-established navigable graphs in a discrete environment provided by the Matterport3D simulator~\cite{MP3D, R2R}. Recent developments have pivoted attention towards continuous environments~\cite{BeyondNavGraph, WaypointModels, DynamicVLN, NavMorph} facilitated by the Habitat simulator~\cite{Habitat1, Habitat2, Habitat3}, where agents engage in fundamental actions such as advancing or turning, thus offering a navigation experience that is more in line with the real-world scenarios.
Transferring discrete learning to continuous environments remains an important challenge~\cite{GridMM}. While previous works attempt to address it using external waypoint predictors~\cite{BridgeGap, VLN-3DFF, li2025ground}, our method resolves this domain discrepancy via instruction-agnostic self-state prediction, which is an intermediate module integrated in the end-to-end pipeline.

\noindent
\textbf{End-to-End Planning.}
Planning serves as the cognitive core of autonomous systems and has been extensively studied in autonomous driving.
Early modular autonomous driving~\cite{VectorNet, MAD-r, xu2023cross} suffer from information loss and error accumulation~\cite{UniAD, VAD}, while subsequent end-to-end solutions~\cite{NVIDIA,AllVehicles} lack interpretability and perform poorly.
Multi-task learning (MTL) attempts to solve the problem by sharing a backbone with separate task heads~\cite{Transfuser,BEVerse, MTL}, but often grapples with negative transfer from isolated optimization targets~\cite{UniAD,MTL,BEVfusion}.
Recently, \textbf{interpretable end-to-end frameworks}~\cite{ST-P3, UniAD, VAD, VLP} integrate perception, prediction, and planning modules into a unified network and significantly improve planning accuracy.
In VLN, the prevailing approaches predominantly fall into either the end-to-end paradigm~\cite{DUET, BEVBert, GridMM} or modular designs~\cite{SpatialNav, FSR-VLN, ImagineNav, ImagineNav2}. In this paper, we adopt the interpretable end-to-end framework to realize our idea, improving the planning performance with intermediate modules.

\section{Methodology}

\begin{figure}[tb]
  \centering
  \includegraphics[width=1\linewidth]{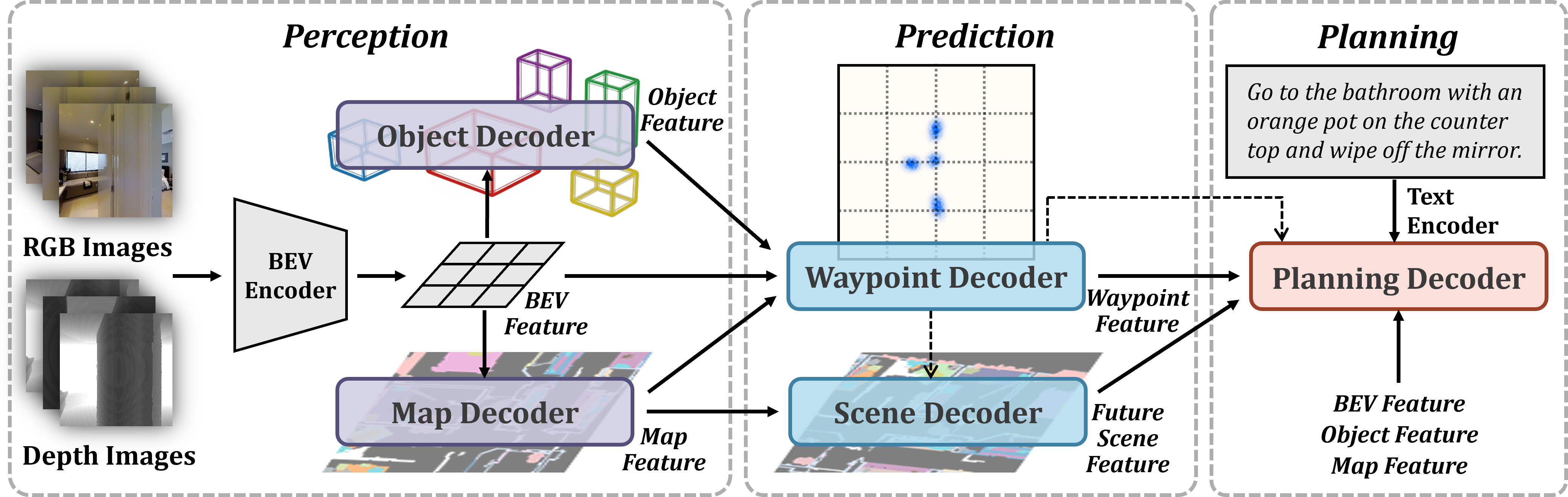}
  \caption{Pipeline of our P$^{3}$Nav model. The agent first encodes its discrete observations into a BEV representation, upon which two perception decoders output object and map features, in parallel. Subsequently, two predictors sequentially decode waypoint cues and future scene semantics. Ultimately, a planning decoder incorporates the features from all the preceding modules to produce a comprehensive navigation decision.
  }
  \label{fig:pipeline}
\end{figure}

As demonstrated in \Cref{fig:pipeline}, our P$^{3}$Nav enhances the standard VLN pipeline with four perception/prediction modules. 
Firstly, our agent encodes its panoramic observations into a unified BEV representation, on top of which the complementary \textbf{object-level and map-level perceptions} are conducted, in parallel (\cref{sec:perception}).
Subsequently, \textbf{waypoint-level and scene-level predictions} are sequentially performed, forecasting the potential future states of the agent and the environment, respectively (\cref{sec:prediction}).
Ultimately, a holistic \textbf{planning} module leverages the features produced by all preceding modules to make a comprehensive navigation decision (\cref{sec:planning}).

\subsection{Perception: Objects and Map Semantics}
\label{sec:perception}
We enhance agent scene understanding with object information and map semantics.
These modules presuppose a unified environmental representation; so we first use Lift-Splat-Shoot (LSS)~\cite{LSS} technology to extract perspective features, lift them into 3D, and project the results into a unified BEV tensor~\cite{ST-P3}. 
To further refine the BEV features, we apply an $L_b$-layer deformable self-attention~\cite{dDETR} encoder upon the LSS outputs, yielding the final BEV representation,
\begin{equation}
\mathbf{B} = \mathtt{SelfAttn}(\mathtt{LSS}(\mathcal{O})) \in \mathbb{R}^{H\times W\times C}, \quad 
\mathcal{O} \triangleq \{O_1,\dots,O_K\},
\end{equation}
where $\mathbf{B}$ denotes BEV feature, $\mathcal{O} \triangleq \{O_1,\dots,O_K\}$ represents the set of multi-view visual observations, and $K$ is the total number of discrete panoramic views.

\noindent
\textbf{Object-level Perception.}
Inspired by~\cite{DETR,DETR3D,BEVFormer2,CLIP-BEVFormer}, we formulate object-level perception as a set-prediction detection task. 
Specifically, $N_{\mathrm{o}}$ object queries 
$\mathbf{Q}_{\mathrm{obj}}\in\mathbb{R}^{N_{\mathrm{o}}\times C}$ are initialized as learnable embeddings. Each query is responsible for 
localizing one potential object by attending to the BEV feature 
$\mathbf{B}$ via a deformable cross-attention transformer decoder,
\begin{equation}
\mathbf{O}=\mathtt{ObjectDecoder}\bigl(
\mathbf{Q}_{\mathrm{obj}},\,\mathbf{B}\bigr) \in \mathbb{R}^{N_{\mathrm{o}}\times C},
\end{equation}
where $\mathbf{O}$ denotes the object features, which are subsequently processed by two parallel multi-layer perceptron (MLP) heads for classification and bounding box regression.
We extract ground truth information from Matterport3D~\cite{MP3D} annotations, and our loss design strictly follows~\cite{BEVFormer}.
The refined object feature $\mathbf{O}$ provides fine-grained landmark cues for downstream cross-modal planning.

\noindent
\textbf{Map-level Perception.}
\begin{figure}[tb]
  \centering
  \includegraphics[width=1\linewidth]{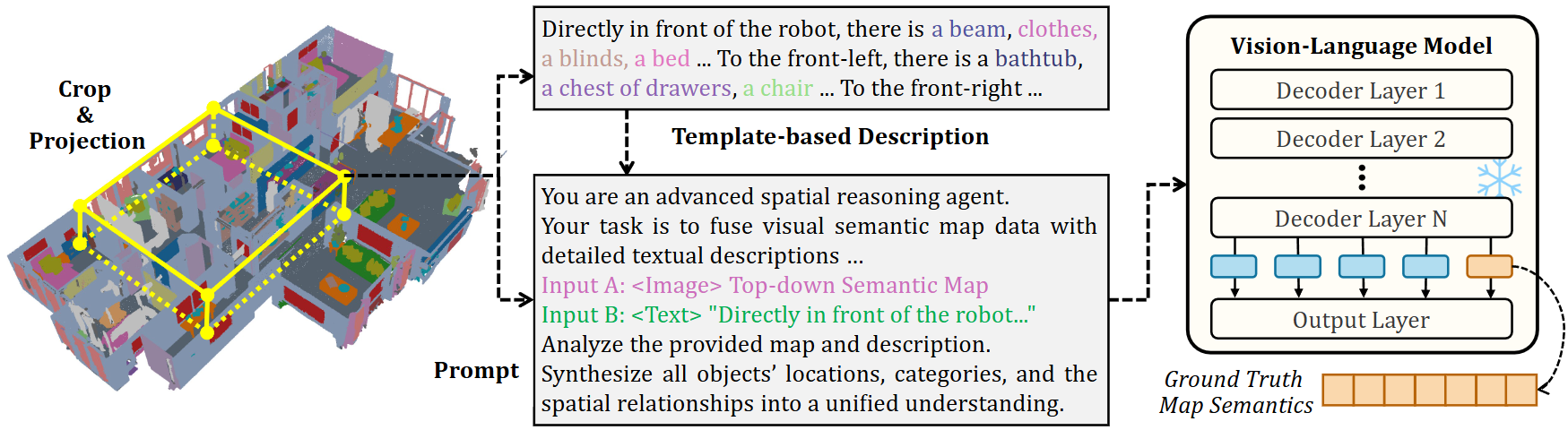}
  \caption{Ground truth map semantics generation. 
  First, we crop nearby regions from the point clouds with semantic annotations in the Matterport3D~\cite{MP3D} dataset, and project them onto a plane.
  Second, we generate a template-based description of the map. Then we prompt the VLM to refine the description.
  The last token from the last VLM decoder layer is defined as the ground truth map semantics.
  }
  \label{fig:mapsemantics}
\end{figure}
To further capture spatial relations among these landmark objects, we perform map-level perception. 
Inspired by~\cite{SegFormer,360BEV}, we initialize a 
learnable query $\mathbf{Q}_{\mathrm{map}}\in\mathbb{R}^{H\times W\times C}$ and let it interact with BEV feature $\mathbf{B}$ through another deformable transformer decoder to acquire map features,
\begin{equation}
\mathbf{M}=\mathtt{MapDecoder}\bigl(\mathbf{Q}_{\mathrm{map}},\,
\mathbf{B}\bigr)  \in \mathbb{R}^{H\times W\times C},
\end{equation}
where $\mathbf{M}$ denotes the decoded map features. 
Since indoor egocentric visual observations are frequently occluded by walls, dense upsampling for pixel-wise map segmentation will introduce a lot of noise. We therefore adopt a compact design: downsampling map features to output a latent code $\mathbf{m}\in\mathbb{R}^{1\times C_m}$ representing map semantics.

The generation pipeline of ground truth map semantics is illustrated in Figure~\ref{fig:mapsemantics}.
We first crop the relevant region from the scene mesh provided by Matterport3D, then project these semantically annotated points onto the top-down plane.
We then generate a template-based textual description for this semantic map. Both the map and the description are fed into a Vision-Language Model (VLM), which fuses these two modalities and refines the description.
Notably, we do not utilize the final textual output of the VLM. Inspired by~\cite{NavGPT2}, we extract the last token $\mathbf{m_\mathrm{vlm}}\in\mathbb{R}^{1\times C_m}$ from the final decoder layer as our map semantics.
We apply MSE loss between the output $\mathbf{m}$ and the ground truth $\mathbf{m_\mathrm{vlm}}$, following~\cite{NavQ}.
The intermediate map feature $\mathbf{M}$ exposes spatial relations in a metrically consistent coordinate frame, strengthening visual-textual association alongside object-level cues.

\subsection{Prediction: Waypoints and Future Scene}
\label{sec:prediction}
\begin{figure}[tb]
  \centering
  \includegraphics[width=1\linewidth]{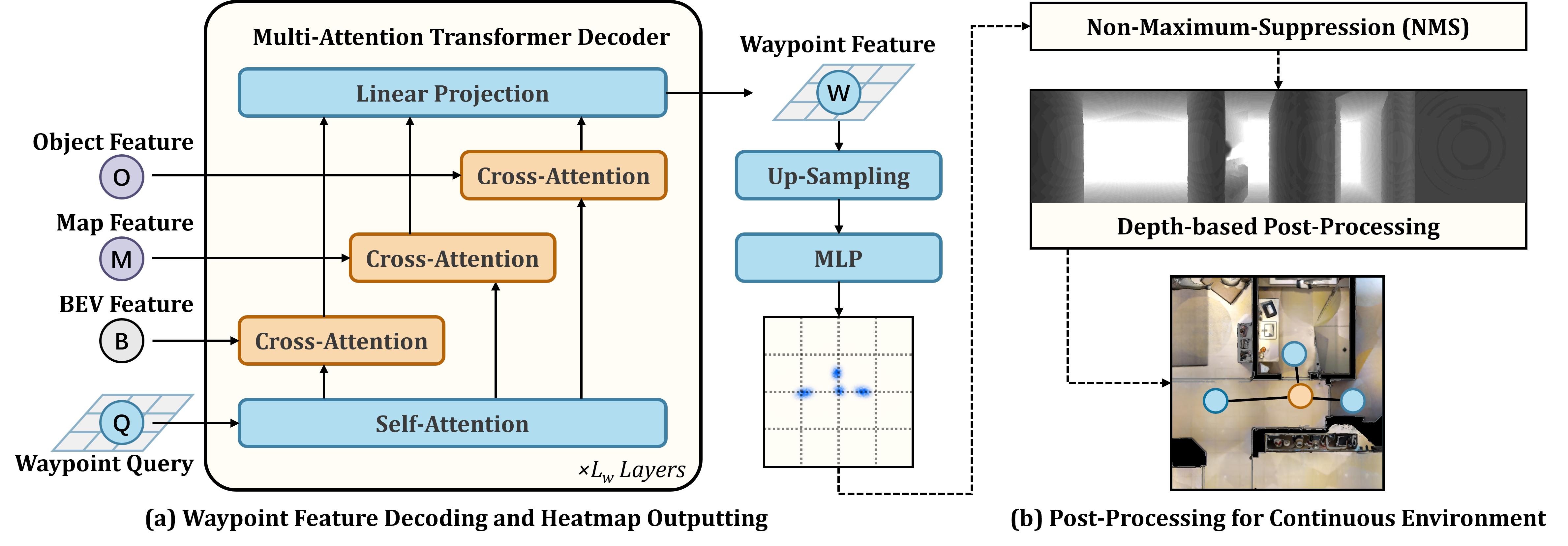}
  \caption{Pipeline of waypoint-level prediction. 
  (a) Waypoint feature is decoded via the multi-attention deformable transformer decoder; after that, a heatmap is generated through up-sampling.
  (b) NMS and depth-based post-processing are performed to identify candidate waypoints, addressing the discrete-to-continuous transfer problem.
  }
  \label{fig:waypoint}
\end{figure}

To equip our P$^{3}$Nav agent with predictive scene understanding capabilities, we employ waypoint-level prediction for the agent's future potential states and scene-level prediction for future environmental information.

\noindent
\textbf{Waypoint-level Prediction.}
Generally, waypoints are viewed as keypoints in the scene where the agent needs to make crucial decisions and are worth exploring~\cite{BridgeGap}, thereby providing a highly adequate representation of the agent's potential future states. 
Predicting these waypoints provides an intrinsic understanding of future locations, a capability critical for informed navigation decisions.

We implement this module via a \textbf{multi-attention transformer decoder} (shown in Figure~\ref{fig:waypoint}(a)).
The decoder updates a learnable query $\mathbf{Q}_{\mathrm{way}}\!\in\!\mathbb{R}^{H\times W\times C}$ by interacting with BEV features $\mathbf{B}$, object features $\mathbf{O}$, and map features $\mathbf{M}$,
\begin{equation}
\mathbf{W}=\mathtt{WaypointDecoder}\bigl(\mathbf{Q}_{\mathrm{way}},\,
[\mathbf{B},\mathbf{O},\mathbf{M}]\bigr) \in\mathbb{R}^{H\times W\times C},
\end{equation}
where $\mathbf{W}$ denotes the decoded waypoint features. 
Specifically, $\mathbf{Q}_{\mathrm{way}}$ first applies self-attention to refine intra-query relationships.  
Subsequently, three parallel cross-attention modules compute queries against $\mathbf{B}$, $\mathbf{O}$, and $\mathbf{M}$, respectively, producing three context-aware tensors.  
These are concatenated with the original $\mathbf{Q}_{\mathrm{way}}$ and fused through a linear layer, yielding a refined output query of identical spatial shape to the raw input.  
The above procedure is repeated for $L_w$ times, giving the final decoded feature $\mathbf{W}$ from our $L_w$-layer multi-attention decoder. 
Next, the waypoint feature $\mathbf{W}$ is upsampled by a stack of transposed convolution layers and finally projected by an MLP head into a pixel-wise heatmap.
Ground truth construction and loss computation follow~\cite{BridgeGap}.
As demonstrated in Figure~\ref{fig:waypoint}(b), we then apply non-maximum-suppression (NMS) and depth-aware filtering to derive candidate waypoints. Note that in discrete environments, waypoints are provided by the simulator during navigation.

\noindent
\textbf{Scene-level Prediction.}
Beyond waypoint-guided self-state forecasting, we further anticipate future environmental semantics surrounding these waypoints, facilitating prospective planning grounded in predictive future context. 
In detail, we leverage the previously constructed map semantics and let our scene decoder directly regress these map-based features.
Technically, we instantiate a learnable future scene query 
$\mathbf{Q}_{\mathrm{scn}}\in\mathbb{R}^{N_{\mathrm{s}}\times C}$ and feed it into the deformable cross-attention decoder with map feature $\mathbf{M}$, akin to the previous design. 
\begin{equation}
\mathbf{S}=\mathtt{SceneDecoder}\bigl(
\mathbf{Q}_{\mathrm{scn}},\,\mathbf{M}\bigr) \in \mathbb{R}^{N_{\mathrm{s}}\times C},
\end{equation}
Critically, the reference points are now waypoint-conditioned, constraining each query to attend regions surrounding actual future map locations. 
As $N_{\mathrm{s}} > P$ where $P$ denotes the number of waypoints, queries unmatched to waypoints are masked, yielding the final future representation $\widehat{\mathbf{S}}\in\mathbb{R}^{P\times C}$.
The loss is computed by MSE loss, as follows~\cite{NavQ}. 
We then build a local graph $\mathcal{G}_f$ as the information interface for planning. Each node represents a candidate waypoint and is characterized by its corresponding predicted future scene feature.

\subsection{Planning}
\label{sec:planning}
The planning serves as a human-like navigational reasoning process in P$^{3}$Nav. Rather than relying solely on immediate visual inputs, our planner evaluates decisions across three cognitive dimensions: 

\noindent
\textbf{Immediate Scene Grounding.}
First,  
our agent evaluates its immediate physical affordance and short-term semantic alignment.
We initialize a query with BEV feature $\mathbf{B}$, and refine it through a multi-attention transformer with object feature $\mathbf{O}$, map feature $\mathbf{M}$, and waypoint feature $\mathbf{W}$. 
Subsequently, cross-attention is applied between the refined query and the instruction feature $\mathbf{T}$, followed by an MLP to output navigation scores,
\begin{equation}
s_{\mathrm{isg}}=\mathtt{MLP}\bigl(\mathtt{CrossAttn}\bigl(\mathtt{MultiAttn}\bigl(\mathtt{Linear}(\mathbf{B}),\,
[\mathbf{O},\mathbf{M},\mathbf{W}]\bigr),\mathbf{T}\bigl)\bigl).
\end{equation}

\noindent
\textbf{Prospective Future Evaluation.}
Since relying exclusively on immediate surroundings makes the agent short-sighted, we task our agent to perform a prospective evaluation.
We process the graph $\mathcal{G}_f$ with the anticipated semantics of candidate waypoints (constructed in Section~\ref{sec:prediction}) and text feature $\mathbf{T}$ through a graph transformer, followed by an MLP to obtain future-aware scores $s_{\mathrm{pfe}}$. This allows the agent to score whether a candidate path leads to a region that matches the subsequent steps of the instruction.
\begin{equation}
s_{\mathrm{pfe}}=\mathtt{MLP}\bigl(\mathtt{GraphCrossAttn}\bigl(\mathcal{G}_f,\mathbf{T}\bigl)\bigl).
\end{equation}

\noindent
\textbf{Global Memory Correction.}
Furthermore, to prevent our agent from getting trapped in local dead ends, we introduce a long-term memory mechanism. Following~\cite{BEVSceneGraph}, we construct a global graph $\mathcal{G}_g$ built upon the historical waypoint predictions. We process $\mathcal{G}_g$ with the textual feature $\mathbf{T}$ through another graph transformer to compute global navigation scores, effectively evaluating all candidate waypoints against the overall instruction,
\begin{equation}
s_{\mathrm{gmc}}=\mathtt{MLP}\bigl(\mathtt{GraphCrossAttn}\bigl(\mathcal{G}_g,\mathbf{T}\bigl)\bigl).
\end{equation}

To yield the final navigational decision, we implement a hierarchical fusion strategy that progressively integrates local perceptions with global constraints.
Specifically, the scores from immediate scene grounding and prospective future evaluation are first fused to form a local decision. 
Subsequently, this local score is further integrated with the global memory correction score to yield the final output, ensuring the navigation is calibrated by long-term historical context,
\begin{equation}
s_{\mathrm{final}}=\mathtt{FusionLayer}\bigl(\mathtt{FusionLayer}\bigl(s_{\mathrm{isg}},s_{\mathrm{pfe}}\bigl), s_{\mathrm{gmc}}\bigl).
\end{equation}

Both fusion layers are implemented following the strategy described in~\cite{BEVBert}.
Finally, the agent navigates to the candidate waypoint with the highest score.

\section{Experiments}
\subsection{Experimental Setup}
\textbf{Datasets and Metrics.}
We evaluate our approach on three datasets, comparing our method with the state-of-the-art approaches introduced in Section~\ref{sec:sota}.
For goal-oriented VLN, we employ the REVERIE dataset~\cite{RvR}, which specifies distant target objects to be interacted with. Notably, REVERIE only provides a discrete environment setup without an official continuous counterpart.
For instruction-following VLN, we adopt R2R-CE~\cite{BeyondNavGraph} and RxR-CE~\cite{RxR-CE}, as these benchmarks offer continuous versions that better align with real-world navigation scenarios. Their discrete counterparts, \ie, R2R~\cite{R2R} and RxR~\cite{RxR}, are also evaluated for a comprehensive comparison, with the results provided in the Appendix.
We adopt standard metrics following prior works~\cite{R2R, RvR, RxR}. Success Rate (SR) and Success weighted by Path Length (SPL) are used across all datasets as primary metrics for navigation success and path efficiency. Additionally, REVERIE employs Remote Grounding Success (RGS) and RGS weighted by Path Length (RGSPL) to evaluate distant object localization; R2R-CE uses Navigation Error (NE) and Oracle SR (OSR); RxR-CE adopts normalized Dynamic Time Warping (nDTW) and Success weighted by nDTW (SDTW) to measure trajectory alignment.

\noindent
\textbf{Training Details.}
We first pre-train P$^3$Nav for 200k iterations with batchsize 12 on 4 RTX4090 GPUs, using the AdamW optimizer with a learning rate of \num{1e-4}. 
To stabilize backbone initialization, only the two perception modules are trained for the first 5k iterations. 
In subsequent iterations, we adopt Masked Language Modeling (MLM) and Single-step Action Prediction (SAP) as auxiliary tasks~\cite{HAMT, BEVBert}; for REVERIE, an additional Object Grounding (OG) task is incorporated~\cite{BEVSceneGraph}. 
At each mini-batch, only one pre-training task is sampled, and the corresponding loss is jointly optimized with the intermediate perception and prediction module losses.
Following established practice~\cite{BEVBert, ETPNav}, we fine-tune the network for sequential action prediction with alternating teacher-forcing and student-forcing strategies. 
All backbone and intermediate module weights are frozen, with updates restricted to the planning decoder parameters. 
This stage runs for 50k iterations with batchsize 8 and a reduced learning rate of \num{1e-5}.

\subsection{Comparison with State-of-the-Art}
\label{sec:sota}
\begin{table*}[t]
\centering
\caption{Comparison with SOTA methods on REVERIE dataset~\cite{RvR}. 
The best and second-best results are marked as \textbf{bold} and \underline{underline}, respectively.
}
\label{tab:reverie}
\begin{tabular}{@{}lcccccccc@{}}
\toprule
& \multicolumn{4}{c}{Validation Unseen} & \multicolumn{4}{c}{Test Unseen} \\
\cmidrule(lr){2-5} \cmidrule(lr){6-9}
Methods & SR$\uparrow$ & SPL$\uparrow$ & RGS$\uparrow$ & RGSPL$\uparrow$ & SR$\uparrow$ & SPL$\uparrow$ & RGS$\uparrow$ & RGSPL$\uparrow$ \\ 
\midrule
HAMT~\cite{HAMT} & 32.95 & 30.20 & 18.92 & 17.28 & 30.40 & 26.67 & 14.88 & 13.08 \\
DUET~\cite{DUET} & 46.98 & 33.73 & 32.15 & 23.03 & 52.51 & 36.06 & 31.88 & 22.06 \\
DSRG~\cite{wang2023dual} & 47.83 & 34.02 & 32.69 & 23.37 & 54.04 & 37.09 & 32.49 & 22.18 \\
GridMM~\cite{GridMM} & 51.37 & 36.47 & 34.57 & 24.56 & 53.13 & 36.60 & 34.87 & 23.45 \\ 
BEVBert~\cite{BEVBert} & 51.78 & 36.37 & 34.71 & 24.44 & 52.81 & 36.41 & 32.06 & 22.09 \\ 
BSG~\cite{BEVSceneGraph} & 52.12 & 35.59 & 35.36 & 24.24 & 56.45 & 38.70 & 33.15 & 22.34 \\ 
VER~\cite{VER} & \textbf{55.98} & \textbf{39.66} & 33.71 & 23.70 & 56.82 & 38.76 & 33.88 & 23.19 \\ 
GOAT~\cite{Wang2024GOAT} & 53.37 & 36.70 & \underline{38.43} & \textbf{26.09} & \underline{57.72} & \underline{40.53} & \underline{38.32} & \textbf{26.70} \\ 
\midrule
\rowcolor{blue!5} \textbf{P$^{3}$Nav (Ours) } & \textbf{55.98} & \underline{36.78} & \textbf{39.94} & \underline{26.03} & \textbf{60.06} & \textbf{40.57} & \textbf{39.75} & \underline{26.56} \\ 
\bottomrule
\end{tabular}
\end{table*}

\begin{table*}[t]
\centering
\caption{Comparison with SOTA methods on Validation Unseen splits of R2R-CE~\cite{BeyondNavGraph} and RxR-CE~\cite{RxR-CE} datasets.
The best and second-best results are marked as \textbf{bold} and \underline{underline}, respectively.
}
\label{tab:result}
\begin{tabular}{@{}lccccccccc@{}}
\toprule
& \multicolumn{4}{c}{R2R-CE} & \multicolumn{5}{c}{RxR-CE} \\
\cmidrule(lr){2-5} \cmidrule(lr){6-10}
Methods & NE$\downarrow$ & OSR$\uparrow$ & {SR}$\uparrow$ & {SPL}$\uparrow$ & NE$\downarrow$ & SR$\uparrow$ & SPL$\uparrow$ & {nDTW}$\uparrow$ & {SDTW}$\uparrow$\\
\midrule
{CMA \cite{BridgeGap} }     & 6.20 & 52 & 41 & 36  & 8.76 & 26.50 & 22.12 & 47.03 & 23.74\\
{VLN$\circlearrowright$BERT \cite{BridgeGap} }     & 5.74 & 53 & 44 & 39  & 8.98 & 27.08 & 22.66 & 46.71 & 24.79 \\
{Reborn \cite{RxR-CE} }     & 5.40 & 57 & 50 & 46 & 5.98 & 48.60 & 42.05 & 63.35 & 41.82 \\
ETPNav \cite{ETPNav}       & 4.71 & 65 & 57 & 49  & 5.64 & 54.79 & 44.89 & 61.90 & 45.33 \\
BEVBert \cite{BEVBert}     & 4.57 & 67 & 59 & 50  & 5.54 & 55.47    & 45.32  & 62.45 & 46.01 \\
HNR-VLN \cite{HNR-VLN}          & \underline{4.42} & 67 & \underline{61} & 51  & 5.51 & 56.39 & 46.73 & 63.56 & 47.24 \\
G3D-LF \cite{g3d}       & 4.53 & \underline{68} & \underline{61} & \textbf{52}  & \underline{5.47} & \underline{57.10} & \underline{47.25} & \underline{63.88} & \underline{47.61} \\
\midrule
\rowcolor{blue!5} \textbf{P$^{3}$Nav (Ours)} & \textbf{4.39} & \textbf{69} & \textbf{62} & \textbf{52} & \textbf{5.42} & \textbf{58.01} & \textbf{47.92} & \textbf{64.29} & \textbf{48.04} \\
\bottomrule
\vspace{-15pt}
\end{tabular}
\end{table*}

\noindent
\textbf{REVERIE.}
As shown in Table~\ref{tab:reverie}, P$^3$Nav achieves new SOTA performance on the goal-oriented REVERIE benchmark. 
In terms of SR and RGS, our method establishes new best results on the test unseen split with $60.06$ SR and $39.75$ RGS, outperforming the previous best model GOAT~\cite{Wang2024GOAT}.
For SPL and RGSPL, our results are comparable to GOAT while significantly surpassing our baseline BEVBert~\cite{BEVBert} by $4.16$ SPL and $4.47$ RGSPL on the test unseen split. 
We attribute these consistent gains to our unified perception and prediction modules, which enhance scene understanding by providing richer and predictive scene features and better associating visual landmarks with target objects specified in the instructions.

\noindent
\textbf{R2R-CE.}
Table~\ref{tab:result} shows that, in instruction-following VLN, P$^3$Nav attains the best results across all metrics on R2R-CE with $4.39$ NE, $69$ OSR, $62$ SR, and $52$ SPL, outperforming the previous best model G3D-LF~\cite{g3d}. Notably, our method surpasses BEVBert~\cite{BEVBert} significantly, demonstrating that scene features extracted by intermediate modules play a vital role in aligning visual information with instruction descriptions.

\noindent
\textbf{RxR-CE.}
On the RxR-CE benchmark (shown in Table~\ref{tab:result}), P$^3$Nav establishes new best results with $5.42$ NE, $58.01$ SR, $47.92$ SPL, $64.29$ nDTW, and $48.04$ SDTW, outperforming the previous best model G3D-LF~\cite{g3d}. These consistent gains across all trajectory alignment metrics validate that our intermediate perception module enables better alignment between current visual landmarks and their corresponding instruction phrases, while the prediction modules allow prospective matching between anticipated visual cues and upcoming language segments.

\subsection{Analysis of Intermediate Modules}

\begin{figure}[tb]
  \centering
  \includegraphics[width=1\linewidth]{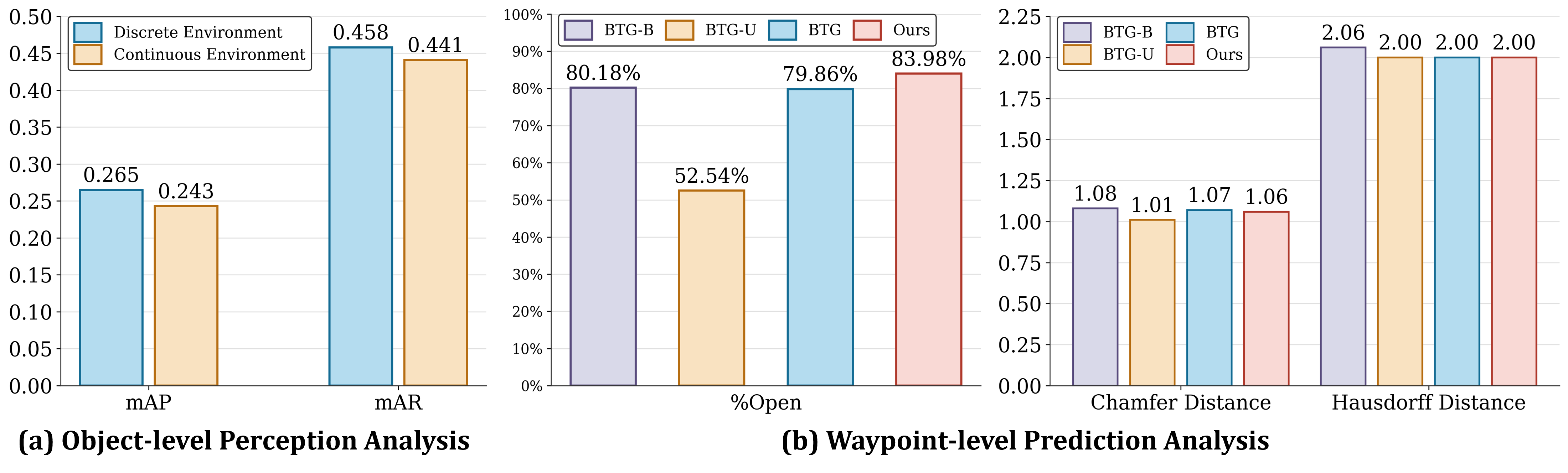}
  \caption{Analysis of (a) object-level perception (mAP and mAR at IoU=0.5) and (b) waypoint-level prediction. BTG denotes~\cite{BridgeGap}, with BTG-B and BTG-U indicating its baseline version and U-Net~\cite{ronneberger2015u} variants, respectively.}
  \label{fig:Quantitative}
\end{figure}

In the previous section, we analyzed the overall navigation performance of our model. In this section, we examine the human-readable outputs of the intermediate modules to provide deeper insights into how perception and prediction contribute to the final planning decisions. 

As shown in Figure~\ref{fig:Quantitative}(a), object-level perception performs slightly better in discrete environments than continuous ones, likely due to rendered observations in continuous environments exhibiting noticeable distortion.
Nevertheless, the overall metrics remain strong, indicating that successful object-level perception constitutes one of the key factors in improving navigation performance.
For waypoint-level prediction (shown in Figure~\ref{fig:Quantitative}(b)), while maintaining comparable Chamfer and Hausdorff distances, our \%Open metric -- measuring the ratio of predicted waypoints located in open space (unhindered by obstacles)\cite{BridgeGap} -- significantly exceeds all prior methods, indicating more practical waypoints for downstream navigation.
We believe this constitutes another contributing factor to our improved navigation performance.

\subsection{Case Studies}
\begin{figure}[tb]
  \centering
  \includegraphics[width=1\linewidth]{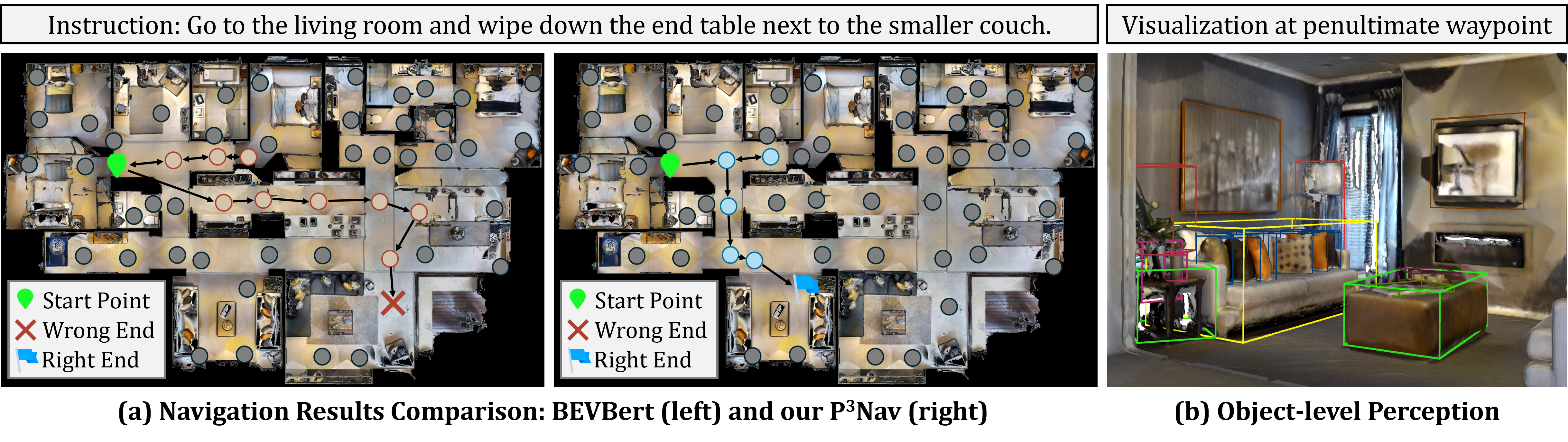}
  \caption{Case study in the simulator. 
  (a) Comparison of navigation results: BEVBert~\cite{BEVBert} (left) fails to find the target, while our P$^{3}$Nav (right) leverages perception and prediction information for accurate waypoint selection.
  (b) Visualization of object-level perception.
  }
  \label{fig:Qualitative}
\end{figure}

\begin{figure}[tb]
  \centering
  \includegraphics[width=1\linewidth]{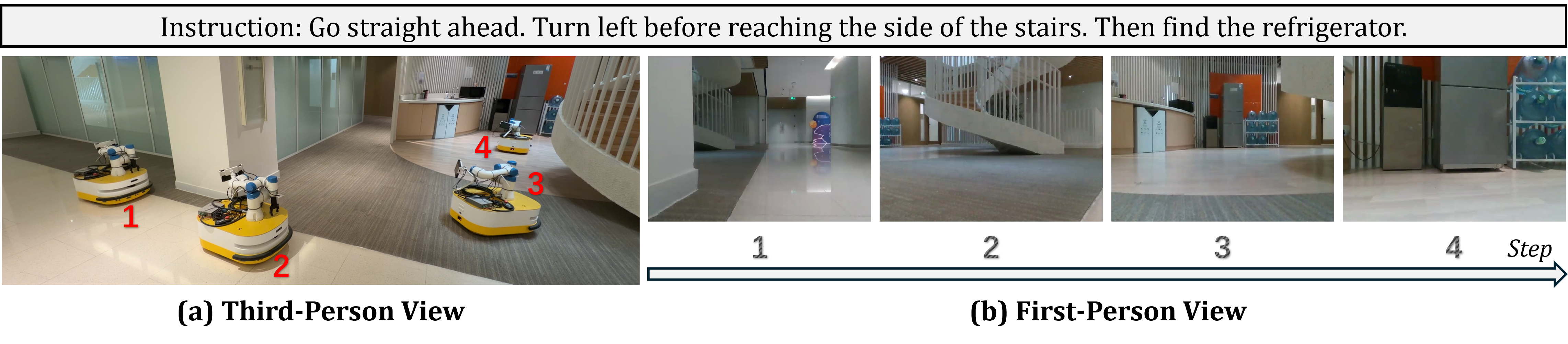}
  \caption{Case study in the real world. 
  (a) Visualization of the path from the third-person view.
  (b) The robot's first-person observations during navigation.
  }
  \label{fig:Qualitative2}
\end{figure}

We visualize the navigation trajectories of BEVBert~\cite{BEVBert} and our P$^{3}$Nav in Figure~\ref{fig:Qualitative}(a). 
BEVBert directly performs planning upon latent BEV features without intermediate modules. 
Due to the absence of a prediction module, BEVBert took more wrong steps forward before triggering backtracking behavior.
Furthermore, without a perception module, BEVBert fails to correctly associate scene information with the linguistic reference \textit{``the \textbf{end table} next to the \textbf{smaller couch}''}, ultimately arriving at an incorrect location.
In contrast, our P$^{3}$Nav leverages its unified perception-prediction-planning pipeline to address these limitations. 
Figure~\ref{fig:Qualitative}(b) shows our P$^{3}$Nav's object-level perception results at the penultimate waypoint, where it correctly identifies the objects mentioned in the instruction.
Additionally, we deploy our algorithm on a real-world robot, providing real-world case study in Figure~\ref{fig:Qualitative2}.
As illustrated, the agent demonstrates precise instruction following and successfully identifies the target object.

\subsection{Ablation Studies}
We conduct extensive ablation studies to validate the effectiveness of each key design. All experiments are performed on the validation unseen split.

\begin{table*}[t]
\centering
\caption{Ablation study of intermediate modules.
When ablating the waypoint decoder, we employ an external waypoint predictor~\cite{BridgeGap} for R2R-CE~\cite{BeyondNavGraph} and RxR-CE~\cite{RxR-CE}.
}
\label{tab:ablation}
\begin{tabular}{@{}l>{\centering\arraybackslash}p{0.9cm}>{\centering\arraybackslash}p{0.9cm}>{\centering\arraybackslash}p{0.9cm}>{\centering\arraybackslash}p{0.9cm}>{\centering\arraybackslash}p{0.9cm}>{\centering\arraybackslash}p{0.9cm}>{\centering\arraybackslash}p{1.2cm}>{\centering\arraybackslash}p{1.2cm}@{}}
\toprule
& \multicolumn{2}{c}{REVERIE} & \multicolumn{2}{c}{R2R-CE} & \multicolumn{4}{c}{RxR-CE} \\
\cmidrule(lr){2-3} \cmidrule(lr){4-5} \cmidrule(lr){6-9}
Model Configs & SR$\uparrow$ & SPL$\uparrow$ & {SR}$\uparrow$ & {SPL}$\uparrow$ & SR$\uparrow$ & SPL$\uparrow$ & {nDTW}$\uparrow$ & {SDTW}$\uparrow$\\
\midrule
Base Model & 50.96 & 36.02 & 59.02 & 50.07 & 55.46 & 45.32 & 62.45 & 45.99 \\
\midrule
\textit{w/o} Object Decoder & 54.13 & 36.58 & 61.79 & 51.89 & 57.45 & 46.67 & 63.38 & 47.53 \\
\textit{w/o} Map Decoder & 53.85 & 36.53 & 60.83 & 50.94 & 56.89 & 46.45 & 63.41 & 47.46 \\
\textit{w/o} {Waypoint Decoder } & 55.04 & 36.67 & 61.87 & 52.09 & 57.47 & 47.71 & 63.78 & 47.85 \\
\textit{w/o} Scene Decoder & 53.68 & 36.17 & 61.22 & 51.45 & 57.14 & 47.10 & 62.98 & 47.06 \\
\midrule
\rowcolor{blue!5} 
\textbf{Full Model} & \textbf{55.98} & \textbf{36.78} & \textbf{62.17} & \textbf{52.32} & \textbf{58.01} & \textbf{47.92} & \textbf{64.29} & \textbf{48.04} \\
\bottomrule
\vspace{-15pt}
\end{tabular}
\end{table*}

\begin{table*}[t]
\centering
\caption{Comparison of different BEV scales.
Odd scales are adopted to guarantee that the agent resides in the center cell, with the cell size fixed at $0.5$m.
}
\label{tab:ablation2}
\begin{tabular}{@{}l>{\centering\arraybackslash}p{1.2cm}>{\centering\arraybackslash}p{1.2cm}>{\centering\arraybackslash}p{1.2cm}>{\centering\arraybackslash}p{1.2cm}>{\centering\arraybackslash}p{1.2cm}>{\centering\arraybackslash}p{1.2cm}>{\centering\arraybackslash}p{1.2cm}>{\centering\arraybackslash}p{1.2cm}@{}}
\toprule
& \multicolumn{2}{c}{REVERIE} & \multicolumn{2}{c}{R2R-CE} & \multicolumn{4}{c}{RxR-CE} \\
\cmidrule(lr){2-3} \cmidrule(lr){4-5} \cmidrule(lr){6-9}
Scale & SR$\uparrow$ & SPL$\uparrow$ & {SR}$\uparrow$ & {SPL}$\uparrow$ & SR$\uparrow$ & SPL$\uparrow$ & {nDTW}$\uparrow$ & {SDTW}$\uparrow$\\
\midrule
{11 $\times$ 11} & 54.91 & 36.52 & 61.78 & 51.99 & 57.61 & 47.81 & 63.97 & 47.85 \\
{15 $\times$ 15} & \textbf{55.98} & \textbf{36.78} & \textbf{62.17} & \textbf{52.32} & \textbf{58.01} & \textbf{47.92} & \textbf{64.29} & \textbf{48.04} \\
{21 $\times$ 21} & 55.73 & 36.69 & 62.01 & 52.18 & 57.95 & 47.88 & 64.22 & 47.99 \\
\bottomrule
\vspace{-15pt}
\end{tabular}
\end{table*}

\begin{table*}[t]
\centering
\caption{Comparison of modular and end-to-end designs for integrating perception and prediction intermediate modules in the navigation pipeline.
}
\label{tab:ablation3}
\begin{tabular}{@{}l>{\centering\arraybackslash}p{1.1cm}>{\centering\arraybackslash}p{1.1cm}>{\centering\arraybackslash}p{1.1cm}>{\centering\arraybackslash}p{1.1cm}>{\centering\arraybackslash}p{1.1cm}>{\centering\arraybackslash}p{1.1cm}>{\centering\arraybackslash}p{1.2cm}>{\centering\arraybackslash}p{1.2cm}@{}}
\toprule
& \multicolumn{2}{c}{REVERIE} & \multicolumn{2}{c}{R2R-CE} & \multicolumn{4}{c}{RxR-CE} \\
\cmidrule(lr){2-3} \cmidrule(lr){4-5} \cmidrule(lr){6-9}
Design & SR$\uparrow$ & SPL$\uparrow$ & {SR}$\uparrow$ & {SPL}$\uparrow$ & SR$\uparrow$ & SPL$\uparrow$ & {nDTW}$\uparrow$ & {SDTW}$\uparrow$\\
\midrule
{Modular} & 54.15 & 36.29 & 61.73 & 51.81 & 57.42 & 47.57 & 63.59 & 47.90 \\
\textbf{End-to-End } & \textbf{55.98} & \textbf{36.78} & \textbf{62.17} & \textbf{52.32} & \textbf{58.01} & \textbf{47.92} & \textbf{64.29} & \textbf{48.04} \\
\bottomrule
\vspace{-15pt}
\end{tabular}
\end{table*}

\begin{table*}[t]
\centering
\caption{Comparison of different ground truth generation methods for map semantics. 
}
\label{tab:ablation4}
\begin{tabular}{@{}>{\centering\arraybackslash}p{1.1cm}>{\centering\arraybackslash}p{1.1cm}>{\centering\arraybackslash}p{1.1cm}>{\centering\arraybackslash}p{1.1cm}>{\centering\arraybackslash}p{1.1cm}>{\centering\arraybackslash}p{1.1cm}>{\centering\arraybackslash}p{1.1cm}>{\centering\arraybackslash}p{1.2cm}>{\centering\arraybackslash}p{1.2cm}@{}}
\toprule
& \multicolumn{2}{c}{REVERIE} & \multicolumn{2}{c}{R2R-CE} & \multicolumn{4}{c}{RxR-CE} \\
\cmidrule(lr){2-3} \cmidrule(lr){4-5} \cmidrule(lr){6-9}
Methods & SR$\uparrow$ & SPL$\uparrow$ & {SR}$\uparrow$ & {SPL}$\uparrow$ & SR$\uparrow$ & SPL$\uparrow$ & {nDTW}$\uparrow$ & {SDTW}$\uparrow$\\
\midrule
{$\mathcal{V}_m$} & 53.82 & 36.51 & 61.32 & 51.44 & 57.02 & 47.13 & 63.53 & 47.66 \\
{$\mathcal{T}$} & 54.46 & 36.59 & 61.41 & 51.56 & 57.21 & 47.25 & 63.60 & 47.72 \\
{$\mathcal{V}_e$} & 55.43 & 36.70 & 61.96 & 52.02 & 57.88 & 47.79 & 64.11 & 47.89 \\
\textbf{$\mathcal{V}_i$} & \textbf{55.98} & \textbf{36.78} & \textbf{62.17} & \textbf{52.32} & \textbf{58.01} & \textbf{47.92} & \textbf{64.29} & \textbf{48.04} \\
\bottomrule
\vspace{-15pt}
\end{tabular}
\end{table*}

\noindent
\textbf{1) Effect of Perception/Prediction Modules.} 
As shown in Table~\ref{tab:ablation}, removing any of the intermediate modules leads to substantial performance degradation across all benchmarks, validating their complementary roles in enhancing scene understanding. 
The object decoder and map decoder jointly strengthen current scene perception and visual-textual alignment: without the object decoder, performance drops noticeably on all datasets, while removing the map decoder causes comparable or even more severe declines, as these modules provide explicit 3D object features and spatial relation cues essential for grounding landmark references in instructions. 
The scene decoder contributes critically to predictive understanding by forecasting semantic map features at future waypoints; its absence significantly harms performance, demonstrating that prospective scene awareness is indispensable for navigation. 
Additionally, the waypoint decoder yields consistent gains, as predicting the agent's own future states fosters intrinsic awareness of navigable locations.

\noindent
\textbf{2) Effect of BEV Scales.} 
As shown in Table~\ref{tab:ablation2}, the $11\times11$ scale underperforms due to limited perceptual scope, while the $21\times21$ scale yields no further gains---distant regions are often wall-occluded and unobservable, forcing perception of invisible areas introduces noise, and long-range dependencies are better handled by global graphs~\cite{BEVBert} rather than enlarged local BEV representation. Thus, $15\times15$ strikes the optimal balance.

\noindent
\textbf{3) End-to-End v.s. Modular Design.} 
We compare our end-to-end pipeline with a modular alternative where perception and prediction modules are independently trained, and their outputs are re-encoded as features for planning. 
As shown in Table~\ref{tab:ablation3}, the end-to-end design consistently outperforms the modular counterpart across all benchmarks and all metrics. 
This validates that our unified pipeline reduces information loss and error accumulation.

\noindent
\textbf{4) Ground Truth Generation for Map Semantics.} 
Table~\ref{tab:ablation4} compares different ground truth generation methods for map semantics. 
Directly encoding semantic maps to visual features ($\mathcal{V}_m$) or using template-based text descriptions ($\mathcal{T}$) underperforms, as they lack cross-modal alignment or semantic refinement. 
While VLM-refined descriptions ($\mathcal{V}_e$) improve performance, extracting the final VLM decoder token ($\mathcal{V}_i$) yields the best results across all benchmarks, demonstrating that latent VLM representations better capture map semantics than all other methods.

\section{Conclusion}
In this paper, we present P$^{3}$Nav, an end-to-end framework that unifies perception, prediction, and planning for vision-and-language navigation.
By perceiving object-level landmarks and map-level spatial relations, our approach significantly enhances the alignment between visual observations and linguistic instructions for navigation-critical scene understanding.
Furthermore, our waypoint-level and scene-level prediction modules enable predictive understanding and planning by leveraging the predicted future states of both the agent itself and the surrounding scene, with map-based future scene features streamlining the planning process by eliminating the need to align discrete view-based embeddings.
Extensive experiments demonstrate that P$^{3}$Nav achieves new state-of-the-art performance on three challenging VLN benchmarks, validating the effectiveness of our unified design. 
We hope this work inspires future research toward more interpretable and holistic end-to-end embodied navigation systems.

%
%
\bibliographystyle{splncs04}
\bibliography{main}

\newpage
\appendix
\noindent\textbf{\Large Appendix}

\section{Problem Formulation}
In Vision-and-Language Navigation (VLN), an agent is tasked to reach a target location specified by a language instruction $\mathcal{W}=\{w_{l}\}_{l=1}^{L}$. In discrete environments, the scene is modeled as a connectivity graph $\mathcal{G}=\{\mathcal{V},\mathcal{E}\}$, where $\mathcal{V}$ denotes navigable nodes and $\mathcal{E}$ represents connectivity edges~\cite{GridMM, BEVBert}. At time step $t$, the agent perceives the environment through a panoramic observation $O_t$, consisting of RGB images $\mathcal{R}_{t}$, depth images $\mathcal{D}_{t}$, and its current pose $P_t$~\cite{BEVBert}. Based on the instruction and observation, the agent learns a policy $\pi(a_t|\mathcal{W}, O_t)$ to select a high-level action $a_t$ from the navigable candidate neighbors provided by the simulator. In continuous environments, the agent moves through 3D meshes without a predefined graph~\cite{GridMM}. Here, the agent positions $\mathcal{P}_{t}$ are continuous, and navigation is performed by predicting reachable waypoints to bridge the gap with discrete planning strategies~\cite{GridMM}.

\section{Evaluation Datasets and Metrics}
\subsection{Datasets}

\noindent
\textbf{REVERIE.}
The REVERIE dataset~\cite{RvR} is a typical goal-oriented VLN benchmark. This dataset is built upon the Matterport3D simulator~\cite{MP3D, R2R} and contains approximately 21.7k instructions covering 4,140 objects, with an average instruction length of about 18 words. The core of REVERIE is to locate a specific remote object in an indoor environment based on a natural language description. The instructions are characterized as high-level and contain ``referring expressions''. Instead of providing specific action steps (\eg, \textit{``turn left''}, \textit{``go straight''}), the instructions directly describe the target object and its room location, such as \textit{``Go to the [room] and find the [object]''}. This requires the model to possess strong semantic understanding and the ability to ground language to visual objects, as the target is usually not visible from the starting position. For instance, the instruction \textit{``Walk to the far end of the kitchen and wait by the dining table.''} task the agent to find the dining table in the kitchen.

\noindent
\textbf{R2R \& R2R-CE.}
The R2R dataset~\cite{R2R} is a foundational instruction-following dataset in the VLN domain constructed on Matterport3D~\cite{MP3D}, consisting of approximately 21.5k natural language instructions paired with human-annotated navigation trajectories, with an average instruction length of 29 words. Its instructions are step-by-step and fine-grained, providing a sequence of low-level directions to guide the agent along a specific trajectory. For instance, \textit{``Turn left and exit the room. Cross the hall to the sitting room. Turn left and enter the bedroom on your left. Wait near the bed.''} describes every step of the navigation process.
R2R-CE~\cite{BeyondNavGraph} reformulates the same instruction–trajectory pairs in a continuous environment, requiring agents to execute low-level continuous control actions such as forward movement and rotation under physics constraints. 

\noindent
\textbf{RxR \& RxR-CE.}
The RxR dataset~\cite{RxR} is another instruction-following benchmark designed to be more challenging and less biased than R2R. 
Built on Matterport3D~\cite{MP3D}, RxR is multilingual, containing approximately 126k instructions in English, Hindi, and Telugu. 
Its instructions are notably longer, averaging 78 words, more descriptive, and feature spatiotemporal alignment.
More importantly, the ground-truth paths corresponding to these instructions are often not the shortest paths leading to the goal, thus further testing the model's instruction-following abilities.
For example, \textit{``Walk past the fridge and out of the doorway and turn left before the bathroom. Then turn left and stop in the room with two white sofas opposite each other. Stop next to the corner table with a plant on it.''}.
RxR-CE~\cite{RxR-CE} brings these complex tasks into the continuous environment.

\subsection{Metrics}

\noindent
\textbf{Success Rate (SR).}
SR measures the proportion of trajectories whose final position lies within a predefined distance threshold $\tau$ (typically 3 meters) from the goal. Let $d_T^i$ denote the shortest-path distance between the agent's final position and the goal for the $i$-th episode. SR is defined as
\begin{equation}
\text{SR} = \frac{1}{N} \sum_{i=1}^{N} \mathbf{1}\left[d_T^i \le \tau\right],
\end{equation}
where $\mathbf{1}[\cdot]$ denotes the indicator function and $N$ is the number of evaluation episodes.

\noindent
\textbf{Success weighted by Path Length (SPL).}
SPL accounts for both task success and path efficiency. Let $l_i$ denote the shortest-path distance from the start to the goal, and $p_i$ denote the actual trajectory length executed by the agent. SPL is defined as
\begin{equation}
\text{SPL} = \frac{1}{N} \sum_{i=1}^{N} \mathbf{1}\left[d_T^i \le \tau\right] \cdot \frac{l_i}{\max(p_i, l_i)}.
\end{equation}
This metric penalizes unnecessarily long trajectories while rewarding successful and efficient navigation.

\noindent
\textbf{Remote Grounding Success (RGS).}
RGS measures whether the target object is visible from the final viewpoint:
\begin{equation}
\text{RGS} = \frac{1}{N} \sum_{i=1}^{N} 
\mathbf{1}\big[\text{target is visible in episode } i].
\end{equation}

\noindent
\textbf{RGS weighted by Path Length (RGSPL).}
RGSPL extends SPL to remote grounding tasks:
\begin{equation}
\text{RGSPL} = \frac{1}{N} \sum_{i=1}^{N} \text{RGS}_i \cdot \frac{l_i}{\max(p_i, l_i)}.
\end{equation}

\noindent
\textbf{Navigation Error (NE).}
NE measures the average shortest-path distance between the agent’s final position and the goal:
\begin{equation}
\text{NE} = \frac{1}{N} \sum_{i=1}^{N} d_T^i.
\end{equation}
This metric reflects the residual navigation error in meters.

\noindent
\textbf{Oracle SR (OSR).}
OSR evaluates whether the agent reaches the goal region at any point along its trajectory. Let $d_t^i$ denote the shortest-path distance to the goal at timestep $t$. OSR is defined as
\begin{equation}
\text{OSR} = \frac{1}{N} \sum_{i=1}^{N} 
\mathbf{1}\big[\min_t d_t^i \le \tau\big].
\end{equation}

\noindent
\textbf{Normalized DTW (nDTW).}
Given a predicted trajectory $P = \{p_1,\dots,p_n\}$ and a reference trajectory $R = \{r_1,\dots,r_m\}$, DTW is defined as
\begin{equation}
\text{DTW(R,P)} = \min_{\gamma} \sum_{(i,j)\in\pi} d(r_i, p_j),
\end{equation}
where $\gamma$ denotes a valid alignment path. The normalized DTW score is computed as
\begin{equation}
\text{nDTW} = 
\exp\left(
-\frac{\text{DTW(R,P)}}{\tau \cdot m}
\right),
\end{equation}
where $\tau$ is the success threshold and $m$ is the reference trajectory length.

\noindent
\textbf{Success weighted by nDTW (SDTW).}
SDTW integrates trajectory similarity with task success:
\begin{equation}
\text{SDTW} = \frac{1}{N} \sum_{i=1}^{N} 
\mathbf{1}\big[d_T^i \le \tau\big]
\cdot \text{nDTW}_i.
\end{equation}

\section{Details of Planning Module}

\begin{figure}[tb]
  \centering
  \includegraphics[width=1\linewidth]{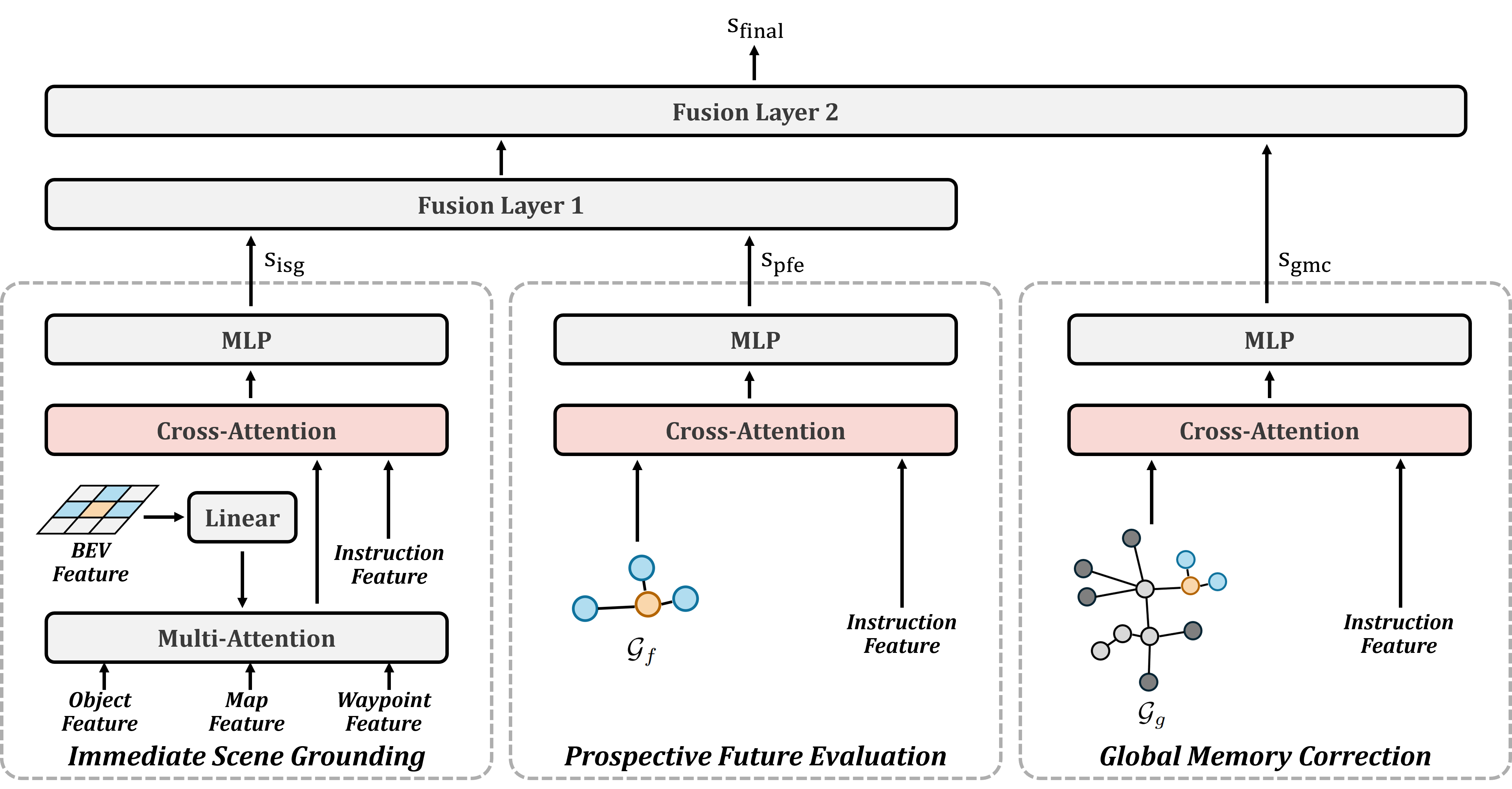}
  \caption{Pipeline of our planning module. Immediate scene grounding evaluates local perceptual feasibility using BEV features and scene semantics.
Prospective future evaluation reasons over candidate waypoint graphs to estimate future consistency with the instruction.
Global memory correction incorporates historical information through a global graph representation.
The resulting scores are progressively integrated through two fusion layers to produce the final navigation score.
  }
  \label{fig:planning}
\end{figure}

The pipeline of our planning module is illustrated in Figure~\ref{fig:planning}.
The planner evaluates candidate waypoints from three complementary perspectives.
First, the \textbf{immediate scene grounding} calculates a navigation score $s_{isg}$ by refining the BEV feature with object, map, and waypoint features, and then attending it to the instruction embeddings.
Second, the \textbf{prospective future evaluation} estimates $s_{pfe}$ by reasoning over the instruction and a local graph that injects future scene features.
Finally, the global navigation score $s_{gmc}$ is computed via \textbf{global memory correction}, which leverages the instruction feature and the global graph constructed during navigation, following~\cite{BEVSceneGraph}.

Subsequently, we adopt a two-stage hierarchical fusion process to integrate these scores, and each fusion layer is designed following prior works~\cite{BEVBert, BEVSceneGraph}.
First, the planner combines the immediate scene grounding score and the prospective future evaluation score to produce a local decision score:

\begin{align}
s_{local} = \alpha\, s_{isg} + (1-\alpha)\, s_{pfe},
\end{align}
where $\alpha$ is a learnable scalar parameter controlling the relative importance of the two signals. 
Following common practice, the weight is constrained to $[0,1]$ via a sigmoid function,

\begin{align}
\alpha = \sigma(w).
\end{align}

Next, the local decision score is further integrated with the global memory correction score to incorporate long-term navigation constraints,

\begin{align}
s_{final} = \beta\, s_{local} + (1-\beta)\, s_{gmc},
\end{align}
where $\beta = \sigma(v)$ is another learnable fusion weight.

After the hierarchical fusion process, the agent selects the candidate waypoint with the highest final score,

\begin{align}
a_t = \arg\max_k s_{final}^{(k)}.
\end{align}

\begin{table*}[t]
\centering
\caption{Comparison with SOTA methods on Validation Unseen splits of R2R~\cite{R2R} and RxR~\cite{RxR} datasets.
The best and second-best results are marked as \textbf{bold} and \underline{underline}, respectively.
}
\label{tab:result}
\begin{tabular}{@{}lccc ccc@{}}
\toprule
& \multicolumn{3}{c}{R2R} & \multicolumn{3}{c}{RxR} \\
\cmidrule(lr){2-4} \cmidrule(lr){5-7}
Methods & NE$\downarrow$ & {SR}$\uparrow$ & {SPL}$\uparrow$ & SR$\uparrow$ & {nDTW}$\uparrow$ & {SDTW}$\uparrow$\\
\midrule
{HAMT \cite{HAMT} }     & 3.65 & 66 & 61  & 56.5 & 63.1 & 48.3\\
{HOP \cite{HOP} }     &  3.80 & 64 & 57  & 42.3 & 52.0 & 33.0 \\
{HOP+ \cite{HOP2} }     & 3.49 & 67 & 61  & 45.7 & 52.0 & 36.0 \\
EnvEdit \cite{li2022envedit}       & 3.24 & 69 & 64  & 62.8 & 68.5 & 54.6 \\
BEVBert \cite{BEVBert}     & 2.81 & 75 & 64  & 68.5 & 69.6 & 58.6 \\
\midrule
\rowcolor{blue!5} \textbf{P$^{3}$Nav (Ours)} & \textbf{2.79} & \textbf{76} & \textbf{65} & \textbf{69.2} & \textbf{70.2} & \textbf{59.1} \\
\bottomrule
\vspace{-15pt}
\end{tabular}
\end{table*}

\section{Additional Experiments}

\subsection{Evaluation on R2R and RxR}
Table~\ref{tab:result} summarizes the comparison between our method and several state-of-the-art methods on the validation unseen splits of both R2R~\cite{R2R} and RxR~\cite{RxR} datasets.
In the R2R benchmark, our P$^3$Nav consistently outperforms all other models across all key metrics. Specifically, we achieve an SR of 76 and an SPL of 65, surpassing the previous best method, BEVBert~\cite{BEVBert}, demonstrating superior precision in localizing target goals in unseen environments.
The advantages of our P$^3$Nav are more pronounced on the more challenging RxR dataset, which contains longer trajectories and more complex linguistic instructions. Our method achieves 69.2 SR and 70.2 nDTW, which are 0.7 and 0.6 absolute improvements over BEVBert, respectively. 
The consistent gains in nDTW and SDTW indicate that our P$^3$Nav is not only more successful in reaching goals but also more faithful to the specific path descriptions provided in the instructions.

\subsection{More Case Studies}
We provide additional case studies in both the simulation and the real world. 

\begin{figure}[tb]
  \centering
  \includegraphics[width=1\linewidth]{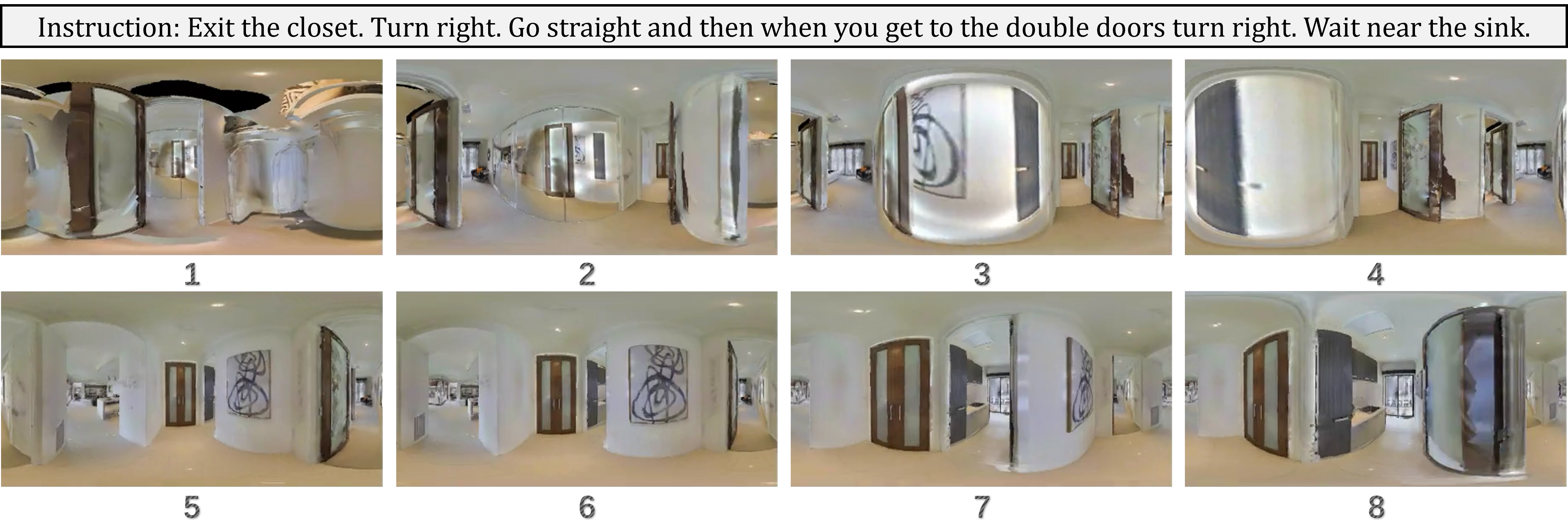}
  \caption{First case study in the simulation. 
  The image sequence shows the robot's panoramic observations during navigation.
  }
  \label{fig:simulation1}
\end{figure}

\begin{figure}[tb]
  \centering
  \includegraphics[width=1\linewidth]{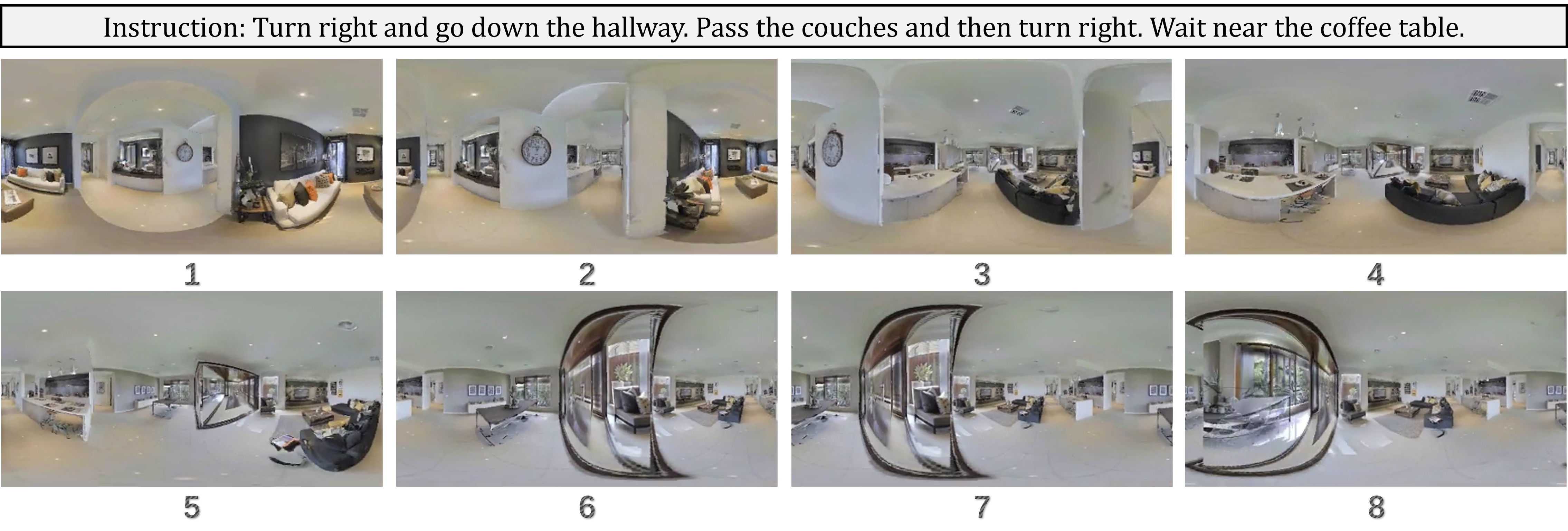}
  \caption{Second case study in the simulation. 
  The image sequence shows the robot's panoramic observations during navigation.
  }
  \label{fig:simulation2}
\end{figure}

\noindent
\textbf{In Habitat Simulator.}
In our first simulation case (illustrated in Figure~\ref{fig:simulation1}), our agent successfully executes a long navigation task. After navigating to exit the `closet', the agent performs an initial right turn and maintains a straight trajectory. Upon encountering the intermediate landmark, it visually grounds the `double doors' and correctly interprets the conditional instruction to `turn right'. Finally, the agent localizes the target `sink' and successfully concludes the episode by stopping `near' it.
In the second simulation case shown in Figure~\ref{fig:simulation2}, the agent initiates the navigation by turning right and advancing down the `hallway'. As it navigates the space, it successfully recognizes the intermediate landmark `couches' and executes the complex spatial action to `pass' them before making a subsequent right turn. Crucially, the agent then shifts its observation to the final destination, successfully localizes the `coffee table', and halts at the required position.

\begin{figure}[tb]
  \centering
  \includegraphics[width=1\linewidth]{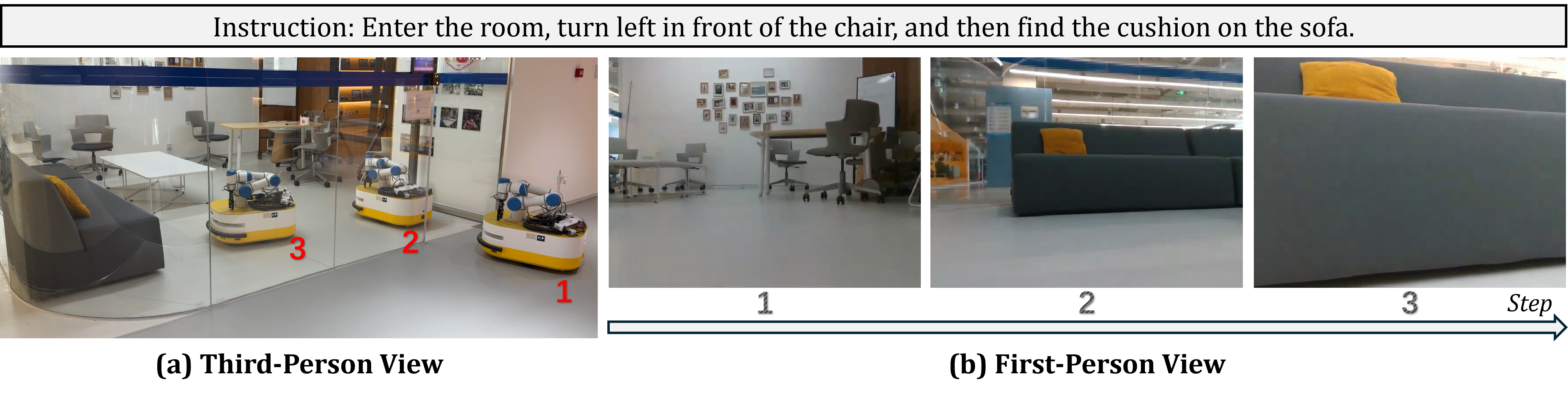}
  \caption{First case study in the real world. 
  (a) Visualization of the path from the third-person view.
  (b) The robot's first-person observations during navigation.
  }
  \label{fig:realworld1}
\end{figure}

\begin{figure}[tb]
  \centering
  \includegraphics[width=1\linewidth]{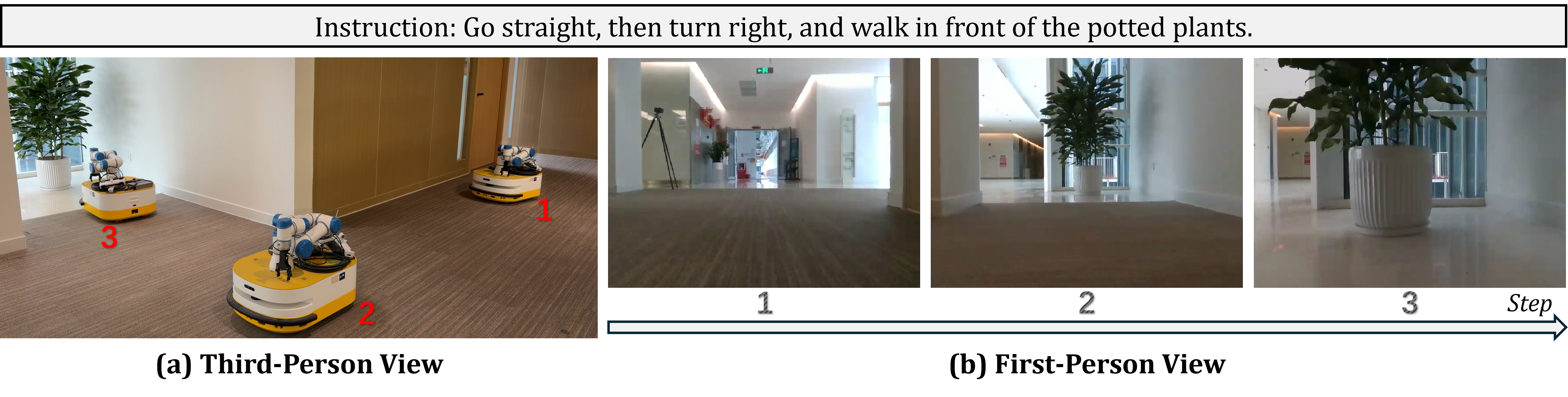}
  \caption{Second case study in the real world. 
  (a) Visualization of the path from the third-person view.
  (b) The robot's first-person observations during navigation.
  }
  \label{fig:realworld2}
\end{figure}

\noindent
\textbf{In Real World.}
In our first real-world case (illustrated in Figure~\ref{fig:realworld1}), our agent successfully executes a multi-step navigation task. Upon entering the room, the agent grounds the semantic landmark `chair' and correctly interprets the spatial constraint `in front of' to perform a timely left turn. Subsequently, the agent shifts its observation to the `sofa' and successfully localizes the target `cushion'. 
In the second real-world case shown in Figure~\ref{fig:realworld2}, the agent first maintains a stable trajectory while executing the `go straight' command. Upon reaching the first waypoint, it successfully identifies the spatial orientation required for a precise `turn right' maneuver. Crucially, the agent recognizes the `potted plants', then adjusts its path to walk `in front of' the target landmark.

Overall, these cases collectively demonstrate our agent's robust capabilities across diverse scenarios: in simulation, it handles long-horizon navigation with complex spatial reasoning and conditional instruction following, while in real-world deployments, it reliably grounds semantic landmarks and adheres to spatial constraints under complex conditions. These consistent successes validate the agent's effective generalization from simulated training to practical, unstructured environments.

\end{document}